\documentclass{bmvc2k}
\runninghead{Tuan Anh \& Dat}{Multi-Stage Attentional U-Net}

\usepackage{enumerate}
\usepackage{amsfonts}
\usepackage{amsmath}

\graphicspath{ {./images/} }

\usepackage[]{color}
\definecolor{cb_orange}{rgb}{1.0,0.51,0.0}
\definecolor{cb_blue}{rgb}{0.22,0.49,0.72}
\definecolor{cb_green}{rgb}{0.3,0.67,0.29}
\definecolor{cb_red}{rgb}{0.89,0.1,0.11}
\definecolor{cb_purple}{rgb}{0.6, 0.31, 0.64}
\title{End-to-End Information Extraction by\\ Character-Level Embedding and Multi-Stage Attentional U-Net
}

\addauthor{Tuan Anh Nguyen Dang}{tadashi@cinnamon.is}{1}
\addauthor{Dat Nguyen Thanh}{marc@cinnamon.is}{1}

\addinstitution{
 Cinnamon AI Labs\\
 Hanoi, Vietnam
}
\addinstitution{
 Cinnamon AI Labs\\
 Hanoi, Vietnam
}

\begin{document}

\maketitle

\begin{abstract}
Information extraction from document images has received a lot of attention recently, due to the need for digitizing a large volume of unstructured documents such as invoices, receipts, bank transfers, etc. 
In this paper, we propose a novel deep learning architecture for end-to-end information extraction on the 2D character-grid embedding of the document, namely the \textit{Multi-Stage Attentional U-Net}. 
To effectively capture the textual and spatial relations between 2D elements, our model leverages a specialized multi-stage encoder-decoders design, in conjunction with efficient uses of the self-attention mechanism and the box convolution.
Experimental results on different datasets show that our model outperforms the baseline U-Net architecture by a large margin while using 40\% fewer parameters. 
Moreover, it also significantly improved the baseline in erroneous OCR and limited training data scenario, thus becomes practical for real-world applications.

\end{abstract}

%-------------------------------------------------------------------------
\section{Introduction}

% \noted{Problem statement}
% The advancement of technology has led to a dramatic change in the way we process data, of which the term "Robotic Process Automation (RPA)" is emerging as an active research domain, and the stemmed results can be used to increase business efficiency, scalability while reducing cost and time. 
% \noted{where "Robotic Process Automation (RPA)" is first introduced? need citation..}
% One of the most important task in conventional RPA works is extracting \noted{is to extract the} structured information from document images, which is a tedious and costly task for humans. \noted{use academic words: time- and man-power consuming. Rule of thumb is to write clear, concise and concrete text, not to deliver long sentences}
% It brings up the development of \noted{It opens up a development branch for}
% document processing systems as a solution to minimize the effort of the human operators in repetitive white-collar works. 
% \noted{Most importantly, missing the brief overview of what information extraction is, you linked RPA to it but did not introduce information extraction and its difficulty, hence weak problem statement.}
%
Information Extraction (IE) is the process of extracting structured information from unstructured documents.
Traditional document processing systems prior to deep learning era often rely on template matching methods to extract information from fixed-format documents \cite{doctemplate_comp_blocks_HanchuanPeng2000, doctemplate_comp_block_prj_Peng2001}. 
However, these methods have scaling issues to operate on a large number of templates, and they are also sensitive to slight input distortions. 
A more general approach to handle varied-format document would be a "document understanding" system, which utilizes a layout analysis step to detect the structure of text-lines, tables and paragraphs in the input image \cite{survey_document_layout_Mao2003, survey_text_extraction_docimage_Ghai2013}. 
After that, an information extraction step \cite{iesurvey_simoes2005, field_extraction_document_images_Rusinol2013} is required to determine their functional roles and semantics.
Designing such a method is considered as a difficult task, especially when the input documents have a high degree of variance and complex layouts \cite{Yang2017}.
Moreover, the errors from preceding processes such as text recognition (OCR) and layout analysis further hinder the performance of various IE methods. 
Hence, there is still much room for improvement to make automatic document extraction become feasible. 
% If the paper has the number of citation large enough, we can cite it as pillars.
% Cite only paper that cited the most

% \noted{Motivation/Solution}
To address the aforementioned problems, we present an end-to-end deep neural network architecture, so-called \textit{Multi-Stage Attentional U-Net}, which can be trained directly with supervision from annotated document data. 
% This formulation consequently eliminates the need for complex hand-crafted heuristics\noted{for example, rule-based? what are they?}. 
Using the 2D-character grid (or char-grid for short) representation of the document as described in \cite{Katti2018}, our network performs pixel-level semantic segmentation task to label and extract the relevant information. 
The char-grid embedding has some substantial differences from natural images, featuring both strong local texture cues and global spatial relationships, thus is challenging for conventional semantic segmentation architecture.   
In order to correctly exploit both textual and spatial features of the char-grid, our network leverages a specialized \textit{multi-stage encoder-decoders} design, where the former encoder-decoder blocks focus on the \textbf{textual components} to identify the important elements (e.g. headers/keywords), and the later ones attempt to learn the \textbf{spatial relations} between elements and propagate that information from the intermediate context to the target values. 
Long-range dependencies and correlations between spatial positions on the document are captured by an additional \textit{self-attention mechanism} \cite{Goodfellow} and the \textit{box convolution layer} \cite{box_conv_Burkov2018}, which further strengthen the ability of the proposed network to model diverse document layouts. 
All modules are differentiable and can be jointly trained in an end-to-end fashion. 
Notably, the training process utilizes a multi-task training scheme with an auxiliary loss and a data augmentation process to enhance the generalization of our model. 
% \lotus{summary: 2-phase U-net to capture textual components and spatial relation, plus self attention and box convolution layer (model design); and auxiliary loss (loss design)}
%
We validate our method on various real-world information extraction datasets and achieve significant improvements over the baseline U-Net model in all tasks while using fewer parameters. 
The network also shows good performance in limited training data scenarios and erroneous results from the OCR step, which matches the standards of industrial needs for automatic document extraction pipeline. 
% \noted{Contribution}
% In summary, we can list our main contributions as follow: \lotus{need revision, reorganized by \@marc and \@tadashi}
In summary, our contribution is several-fold: 

\begin{itemize}
\item We proposed a specialized deep neural network architecture to perform the information extraction task on the 2D character-grid, which largely surpasses the strong baseline U-Net and has 40\% fewer parameters. 
\item We systematically evaluate the use of self-attention mechanism, box-convolution and multi-stage encoder-decoders upon modeling complex spatial relations in the document layout.
Extensive ablation studies are conducted to verify the effectiveness of each component. 
\item Rigorous experiments are performed on our in-house customer-provided datasets to validate the robustness of our method under harsh conditions (limited training data, heavy OCR errors).
\end{itemize}

% \noted{Paper structure}
% The rest of this paper is organized into 3 sections. We first provide related work in \textit{Section \ref{sec:related_work}}, then describe the character embedding presentation of a document in \textit{Section III.A}. In \textit{Section III.B}, we present the details about our proposed architecture for the information extraction task. Finally, we evaluate the results of our network on several datasets quantitatively and qualitatively in \textit{Section IV}.

\section{Related works}
\label{sec:related_work}
% Regarding the problem of information extraction from document image, there have been three main branches of research \lotus{research directions}: 
Information extraction techniques for document images can be roughly classified into three groups: 
the template matching-based methods, the heuristic-based methods and the deep-learning-based methods. 
Those state-of-the-art approaches are reviewed as follows:

\textbf{Template Matching Based Methods:} This research direction formulates the problem of information extraction as document registration~ \cite{doctemplate_comp_blocks_HanchuanPeng2000,doctemplate_comp_block_prj_Peng2001} or form classification \cite{Aldavert2017}, in which image would first be matched with a type of form in a database and the information will be extracted from predefined areas. 
% For extensive information on the template matching methods for document image information extraction, please see \cite{doctemplate_comp_blocks_HanchuanPeng2000, doctemplate_comp_block_prj_Peng2001}. 
Further readings on document image information extraction that uses template matching methods can be referred in \cite{doctemplate_comp_blocks_HanchuanPeng2000,doctemplate_comp_block_prj_Peng2001}.
% Template matching methods have major drawbacks when it comes to undefined document structure form, which limits its application in some document processing pipelines.
However, these methods, in general, have certain drawbacks while dealing with undefined and unseen structured form, which hinder wide-adaptation in automatic document processing pipelines.

\textbf{Heuristic Methods:} The conventional information extracting pipeline \cite{Bukhari2011, Namboodiri2007} often tries to build up the components using carefully designed feature engineering. 
%: \cite{Bukhari2011} proposed a method to detect text line using ridges and white space based on segmentation, \cite{Ahn2017} instead used a technique so-called binarization followed by a step of connected component analysis. 
% The lower level components (i.e text lines) are then used for OCR, to form or classify them to higher-level components such as Title, Paragraph, etc. 
% The higher-level labeled regions are then used to determine the reading orders or to extract the desired information.
%
% Other approaches like \cite{Kise1999} handling the segmentation top-down using either KMeans or Voronoi Diagram. 
Kise \textit{et al.} in their seminal work \cite{Kise1999} introduced a selective method based upon the edges of primary and residual Voronoi graphs that can reduce the cost of character-wise segmentation, which mainly relies on connected component labeling of binary inputs. 
% Development of heuristic segmentation method leads applicability of middle-level heuristic approach: 
As conventional computer vision-based segmentation methods evolved, their adaptations to process image document were exploited:  
while \cite{Chakrabarti2008}, \cite{Oro2009} and \cite{Co2014} used text lines to detect tables or segment page in to regions, \cite{Wang2004} defines table reading orders and heuristically performs table understanding on segmented text lines and separators. 
One major limitation of the heuristic approaches is that they assume a certain condition on the dataset, which limits their generality to be used with different types of forms.

\textbf{Deep learning based methods} Instead of using hand-crafted feature design, 
recent development in deep learning on image segmentation like Fully Convolutional Networks \cite{Long2015} and U-Net \cite{Ronneberger2015} have allowed early attempts in training end to end text line-level segmentation and labeling like that of \cite{Oliveira2018} which applied fully convolutional network to directly extract regions of interest from historical documents. 
% However, due to original limitation of CNN receptive field mechanism as \cite{Yang2017} the methods cannot be applied to small object level, leading to restricted applicability.
Our proposed method is more aligned with recent end-to-end systems which try to incorporate both semantic structure and spatial image features \cite{Yang2017, Palm, Katti2018}. 
While \cite{Yang2017} use sentence-level embedding which assumes perfect text line grouping to segment the document semantically and targets to conventional documents, we aim for information extraction from visually rich documents, of which character-level segmentation is essential. 
% Our major difference in comparison with \cite{Palm} is that we use only an end-to-end structure without using an external memory or requires different model for each interested fields, as well as significantly less data required to train the model. 
% Our method of representing the input is similar to that of \cite{Katti2018} though we also adapted the color representation features, the major different between our method and \cite{Katti2018} lies in the internal processing structures.

% On the final remarks of this section, we briefly describe our model which 
To the best of our knowledge, the proposed neural network-based model
is essentially different from the two aforementioned approaches: since we are considered the problems of progressive information extraction (i.e. first the `key' is segmented before the whole `key-value' is segmented), the potential of multiple stacked U-Nets has to be considered, for which we adapted the structure of the recent coupled U-Net~\cite{Tang2018}, while \cite{Yang2017, Palm, Katti2018} leverage variants of a single-output fully convolutional neural networks. 
Further, the model is adapted with global information propagation capability by the use of self-attention mechanism on each convolutional block, it is noted that box convolution \cite{box_conv_Burkov2018} is also leveraged to support distanced region information incorporation.

Regarding the embedding, the char-grid~\cite{Palm} is chosen and tweaked over the hierarchical levels of sub-word \cite{Bojanowski2016}, word \cite{Mikolov2013, Pennington2015} or sentence level embedding which have been used in \cite{Yang2017}, because, as observed, we note that there are cases like numbers, formatted date, etc. that cannot appropriately be represented using a finite number of vectors nor be modeled accurately by N-grams. 
% Using character-level embedding, the aforementioned architecture can easily be trained using augmented data to achieve flexibility and robustness to noises from line-segmentation, OCR stages.

% \section{Background}
% \input{background}

\section{Method}
\subsection{Chargrid Conversion}

% Our proposed method for information extraction receives input from the text segmentation and OCR process, which consists of the document image $I$ and the set of text boxes $L$:
% \begin{itemize}
%  \item A document image $I \in {\rm I\!R}\left[0, 1\right]^{H \times W \times C}$ where $H$ and $W$ is the height and width of the document image respectively, $C$ is number of channels (i.e. for RGB-colored document image, $C=3$).
%  \item A set of $N$ text line boxes in the document image $L=\{(t_1,b_1),(t_2,b_2),\ldots,(t_N,b_N)\}$ where $t_i$ is the text content of the $i$-th line and $b_i\in {\rm I\!R}[0,1]^4$ denotes the normalized line position and size. 
%  This is a typical output of the text segmentation and OCR modules in a document processing pipeline. 

%  \end{itemize}
% The input is then converted to chargrid format as follow:
The document image, after being processed by a standalone OCR module to form the text lines, can help to construct the chargrid as follows:
\begin{itemize}
    \item \textbf{Character separation:} We first convert the text line boxes to character boxes by dividing them horizontally to the number of character reside in the text with the assumption that in each text line, the variance of width and height of each character are minimal. 
    After this step, a list of character boxes $L_c=\{(c_1, b_1),(c_2, b_2),\ldots,(c_N, b_N)\}$ is obtained where $c_i$ is the character index and $b_i$ its respective bounding box.
    
    \item \textbf{Character encoding:} Each character is indexed, a constant number of most frequent characters $N_{char}$ is chosen. 
    The character boxes are then used to create a mask $CM \in \{0, 1, \dots, N_{char}\}^{H \times W}$.
    For each character box in the list with the character index $i_{char} \in \{1, \dots, N_{char}\}$, the region of the mask covered by the box will be filled with the value $i_{char}$. 
    %\item All remaining character-pixels corresponding to empty regions in the original document page are initialized with $0$. 
    %\item 
    $CM$ is then converted to one-hot format $CM_{one\_hot} \in \{0, 1\}^{H \times W \times (N_{char} +1)}$ 
    %as follows: $CM_{one\_hot}$ is initialized with 0. 
    %For each index $i \in \{0, \dots, H-1\}$ and $j \in \{0, \dots, W-1\}$, $CM_{one\_hot}(i, j, CM(i,j)) = 1$. 
    %The reason behind this conversion is to avoid the case of different performance corresponding to different indexing of characters, leaving the task of compact spatial and textual representation learning of each character to the model.
\end{itemize}

% For the precision required (often at character level), the problem of information extraction on our original input is now reformulated as segmentation on the chargrid:
% Taking $CM_{onehot}$ as input, we wish to train the neural network that can output the mask $M \in \{0, \dots, K-1\}^{H \times W}$ masking the regions of $K$ target fields, with the supervision from labeled mask $LM \in \{0, \dots, K-1\}^{H \times W}$ for each input document.

At this point, the problem of information extraction is now equivalent to the pixel-wise segmentation on the chargrid:
Taking $CM_{one\_hot}$ as input, we wish to train the neural network that can output the mask $M \in \{0, \dots, K-1\}^{H \times W}$ masking the regions of $K$ target fields, with the supervision from labeled mask $LM \in \{0, \dots, K-1\}^{H \times W}$ for each input document.

% $V=\{v_1,v_2,\ldots,v_K\}$ for each input document. 
% The label $V$ has the same format as the target output, where $v_i$ is the set of text boxes belong to the target field $i$-th.
% For the precision required (often at character level), we formulated the problem as semantic segmentation task on document image with additional information of text-boxes. 
% In the first step, the character grid $g$ \cite{Katti2018} is built from the input, which has the same size as the document image $I$.
% For each character $c_k$ at location $b_k$, the area covered by that character is filled with some constant index value $E(c_k)$. 
% The network architecture presented below is then used to segment regions of interest on $g$.

\subsection{Network architecture}

\begin{figure*}[h]
\centering
\includegraphics[width=1\textwidth]{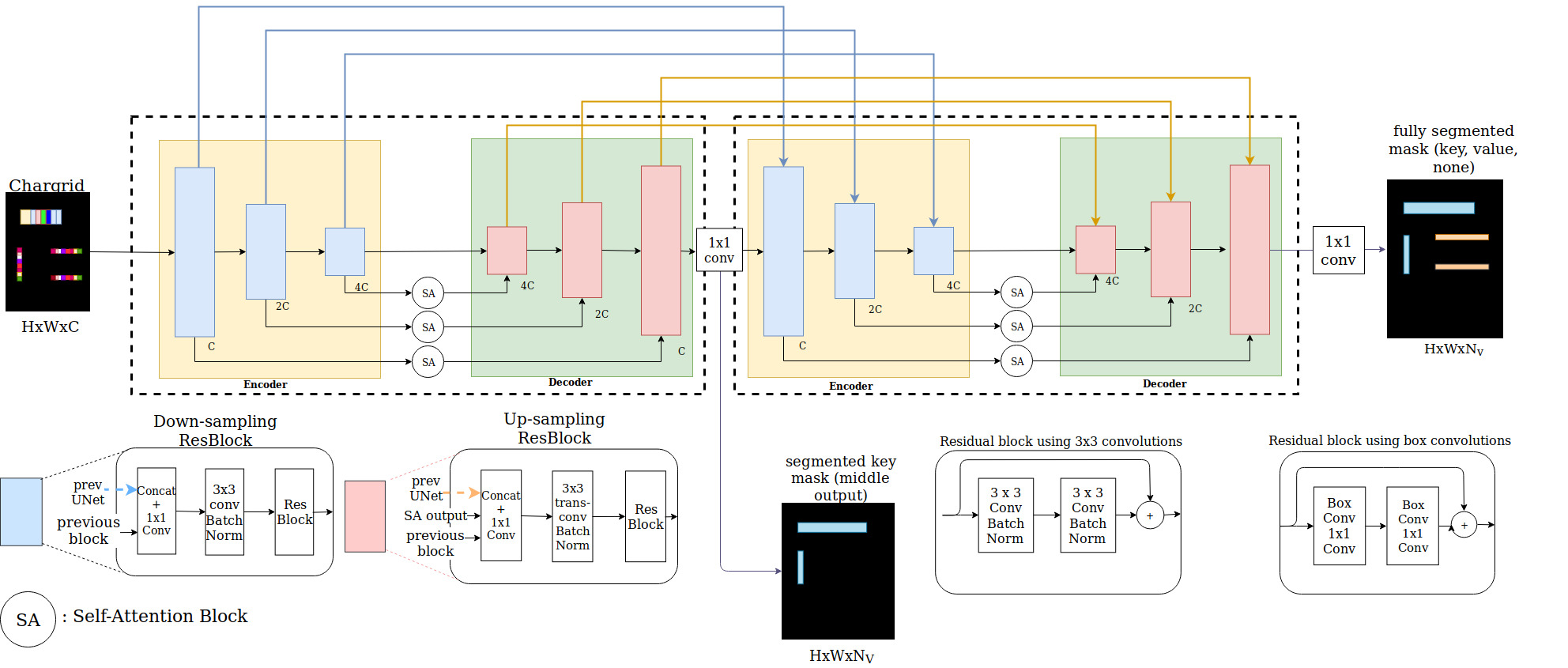}
\caption{Architecture of the \textit{Multi-Stage Attentional U-Net}. \small{The input chargrid goes through several of encoder-decoder blocks to generate the output mask. Both encoder-decoder blocks are tightly connected by the additional coupling-connections. An auxiliary loss is used during training to increase the context-awareness of the model.}}
\label{fig:model_architecture_full}
\end{figure*}

% Unlike articles and books which reading order can be easily determined since the reading order is conventional, extracting information from business documents relies heavily on both textual and spatial information. 
% Hence, our goal is to build a model that can leverage both those kinds of information simultaneously. 

\begin{figure}[h]
    \centering
    \subfigure[The need for spatial information flow in modeling the document semantic layout]{
            \label{fig:propagation_illustration}
            \includegraphics[width=0.45\textwidth]{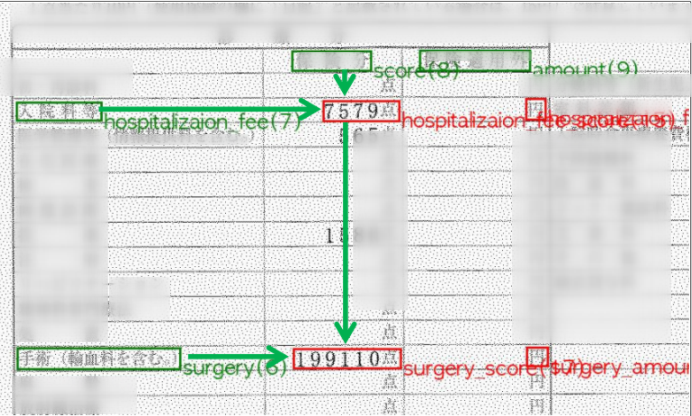}
    }
    ~
    \subfigure[Box convolution operation allowing storing the mean value of distanced region. The output of the stored pixel value is mean of the values of pixels covered by the box]{
        \label{fig:box_convolution}
        \includegraphics[width=0.50\textwidth]{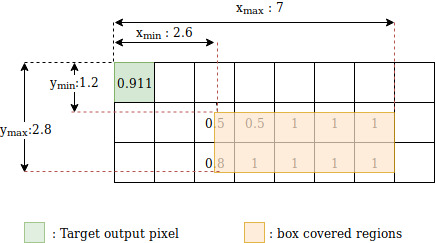}
    }
    \caption{Illustration of spatial information propagation}
\end{figure}

The architecture of our model is depicted in Figure \ref{fig:model_architecture_full}. 
We use the backbone as \textbf{Coupled U-Net}~\cite{Tang2018}, with the inspiration behind this utilization is the benefit of \textbf{multi-stage outputs}. %, making the analysis of model inference straightforward.
For example, in order to understand what kind of information a value in a table represents, a person have to look at the top header or the side description (see Figure~\ref{fig:propagation_illustration}). This order of behavior can be satisfied by forcing the middle U-Net block to segment key and/or description while the final end outputs the full segmentation of all key/value collections. 
%while the final is required to fully segment all types of key/value available.
%The constrains is put on the middle output as follows:
% Specifically, 
% the middle output mask and the auxiliary target mask $AM$, $ALM \in \{0, \dots, K-1\}^{H \times W}$ is of the same dimension as the final output mask $M$ and $LM$. 
% However, the auxiliary target mask constrains is altered: only the key regions segmentation is counted to the loss calculation of middle layer.
%
In addition, other advantages of the Coupled U-Net are also leveraged such as the avoidance of gradient vanishing while promoting feature reuse by the dense skip-connections between corresponding encoder and decoder blocks of each U-Net to another; 
and the use of auxiliary loss applied to the output of middle U-Net block.

\begin{figure}[h]
\centering
\includegraphics[width=0.85\textwidth]{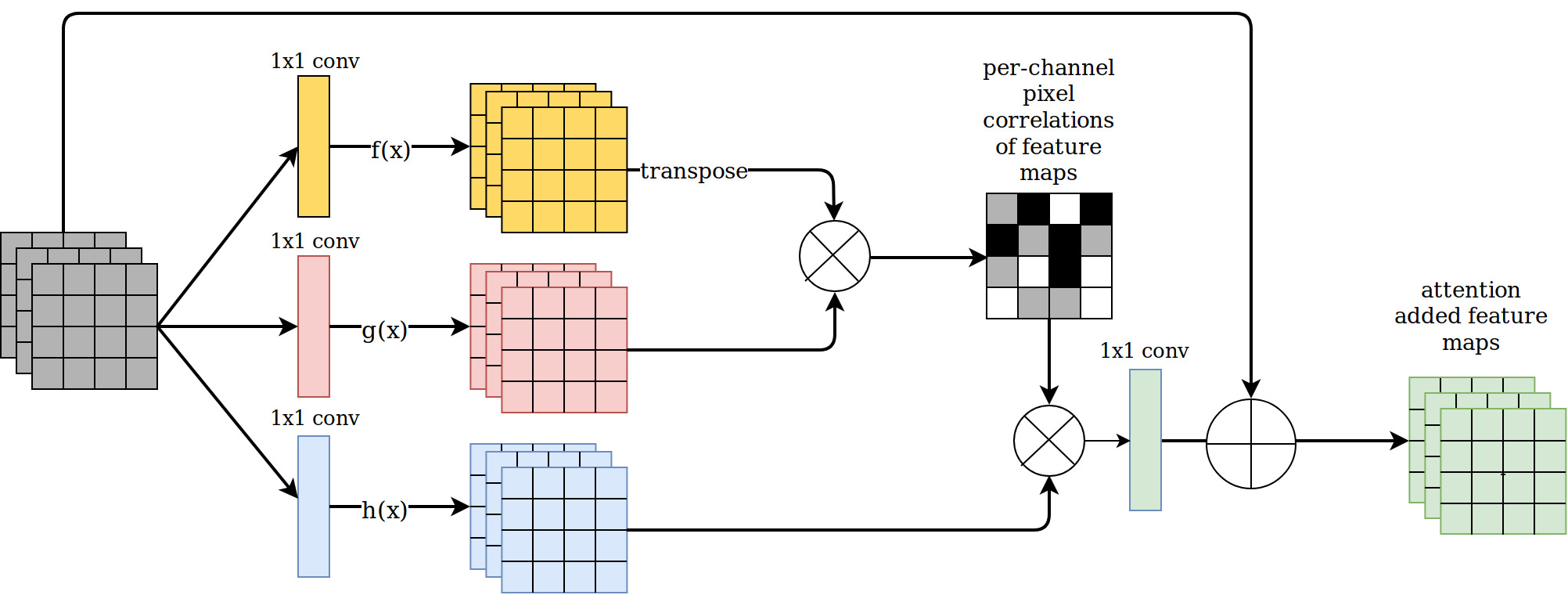}
\caption{The \textit{self-attention mechanism} applied on feature maps. \small{Each pixel's feature is projected to two separate latent spaces by two linear transformation $f, g$ respectively. The per-pixel correlation is then calculated, of which the result is used to aggregate information from each high-correlating pixels to the destination in the output map of the third linear transformation $h$. The propagation of information is cancelled when either the correlation matrix is a zero matrix or the third linear transformation is a mapping to zero.}}
\label{fig:self_attention}

\end{figure}

% The model is further adapted as follows: 
The detail is further explained as follows: 
\textit{Firstly}, for the sake of flexibility, each encoder/decoder block is constructed by chaining multiple \textbf{ResBlocks}, each of which is a ResNet-like bottleneck, with ($3 \times 3$) convolutions, batch normalization and residual connection. 
The downsampling blocks are equipped with dilated convolution while the upsampling blocks use transposed convolution blocks for upscaling their resolutions. 
\textit{Secondly}, from an intuitive observation that different information flows from table headers and descriptors are required to extract the right target value (illustrated in Figure \ref{fig:propagation_illustration}), we incorporate the \textbf{Box-Convolution layer} in \cite{box_conv_Burkov2018} (Figure \ref{fig:box_convolution}) to a variant of ResBlock, which helps modeling long range vertical and horizontal interaction. 
%still inspired by the idea of information propagation from header/description (the key) to identify the types of values,  
% To enable this kind of interaction, we incorporate the \textbf{Box-Convolution layer} in a variant of ResBlock. 
The box convolutions \cite{box_conv_Burkov2018} are sliding mean window with learnable offsets instead of weights over windows (see Figure \ref{fig:box_convolution}). 
Learning the offset patterns aids the model in aggregating information from distant regions in a straight forward way,
making it a suitable alternative to conventional convolution blocks.
\textit{Thirdly}, to strengthen the Coupled U-Net design, expansion paths (also known as \textit{skip-connection}) from the same level of downsampling blocks to upsampling blocks are added, with \textbf{self-attention mechanism} (non-local network) \cite{Goodfellow} (see Figure \ref{fig:self_attention}) to promote global information propagation between each encoder-decoder pair of an U-Net block.

In short, with the described building blocks, we have built an architecture that is light-weight and feasible to exploit both textual and spatial relations from the chargrid embedding.

% \textbf{Self-Attention mechanism}. Spatial context is very crucial in understanding semantic layout of a document, which requires the model to capture the long-range dependencies between objects. However, most models for semantic segmentation are built using convolutional layers, which operate in a local neighborhood. To solve this problem, we adopt the self-attention mechanism in \cite{Goodfellow} to model the spatial relationship between any two points in the feature maps \ref{fig:self_attention}. The self-attention is added to the skip-connection between the encoder and the decoder of an U-Net block to exploit the high-level features from the encoders. We also adapt he positional embedding as described in \cite{Goodfellow} to ensure the network can aware of relative position between pixels. 

\subsection{Training specification}
\label{subsec:training}
We specify the additional detail about the training stage, including model hyper-parameters, the loss function and the data augmentation process. 
The hyper-parameters of the standard $msau$ model is chosen as follows: the number of most-frequent characters $N_{char}=256$, number of U-Net blocks $n_{blocks}=2$, the depth (number of convolution operations) of the ResBlock $res\_depth=2$, the number of channels of the first input feature map $C=16$, the number of downsampling blocks  in an encoder $n_{downsampling}=4$. 
$msau\_big$ has the same configuration except $n_{downsampling}=5$.
The same setups are also used for our baseline variants of U-Net, which contain a single encoder-decoder block.
Our code, trained model and test samples will be tentatively released via this link\footnote{\url{https://github.com/datvo06/MSAU}} to foster the reproducibility of this work. 

\textbf{Loss function}. The multi-task loss used in the training process is a weighted sum of two smaller losses: the ending loss of the final output and the auxiliary loss of the intermediate output. 
We adopt the Focal Loss \cite{focal_loss_Lin2017a} to resolve the class imbalance problem during training. 
The auxiliary loss is computed from the ground-truth keys mask, after removing all the values. The corresponded weight are $\gamma=0.4$ for the auxiliary loss, and $1-\gamma$ is applied to the ending loss.

\textbf{Data augmentation}. In order to increase the robustness of the model against varied layouts, we perform various types of data augmentation. 
The text-boxes in the input document are re-scaled so that the median text height is 3 pixels in the char-grid. 
We use the following steps for augmentation: random character replacement to mimic OCR errors, random text box shifting, random affine transformation, random background padding.
\section{Experiments}
We evaluate and analyze our approach on two customer-provided documents: self-collected \textit{Japanese invoices} and the \textit{medical receipts} datasets. 

\subsection{Datasets}

% Due to the lack of publicly available samples for document information extraction, we collected two industrial datasets for our experiment. 
% They both come from real business requirements and comprise of challenging document images with a variety of layouts. 
Our in-house datasets comprise of challenging document images with a variety of layouts.
The images are also corrupted by several distortions such as noises, blurs, missing strokes due to the scanning process or poor paper quality.
These conditions often appear in a practical document processing pipeline, thus make our datasets a good benchmark for information extraction methods. 
Ground-truth for the data was annotated manually by human operators, which includes the location and content of text-boxes as well as the key information boxes that need to be extracted. 

\textbf{Japanese invoices dataset} consists of 261 invoices in Japanese from several vendors. Each invoice contains 16 key-value pairs to extract. 
Example keys are \textit{date\_issue, sender\_name, receiver\_name, total\_amount, tax, item\_name, item\_amount}, etc. 
%A few sample images are shown in the supplement (Figure \ref{fig:toyota_dataset}).
A few sample images are shown in the supplement (Figure 5).
This dataset features some notable challenges such as: mixed handwriting and printed text, random element placements, low resolution scanning. 

\textbf{Insurance medical receipts dataset} consists of 200 medical receipts in Japanese with varied formats. 
Each document has 12 keys to be extracted: \textit{date\_issue, billing\_period, insurance\_amount, patient\_co-payment\_ratio, surgery\_fee, hospitalization\_fee}, etc. 
%Examples of this dataset can be re-aced via the supplement as well (see Figure \ref{fig:Insurance_dataset}).
Examples of this dataset can be accessed via the supplement as well (see Figure 6).
The challenges of this data include long-range key-value correlation, multiple keys relation, skewed images. 

Both datasets are divided into 70\% for training and 30\% for testing.  

\subsection{Implementation details}

The proposed network was implemented using Tensorflow library. 
We perform experiments on a server equipped by an Nvidia Quadro M4000 (8GB memory GPU). 
All models were trained for 200 epochs with a mini-batch size of 4. 
The \textit{RMSProp} method \cite{rmsprop_tieleman2012} is used for optimizing the model. 
The learning rate is set to $0.001$ at the beginning and decreases every 10 epochs with a polynomial decay of 0.9.

% The standard Multi-stage Attentional U-Net ($msau$) is implemented with the number of convolution layers in a residual block $res \_depth = 2$ and the base feature depth $C=16$. Each encoder/decoder has 4 residual/upsampling blocks respectively. The $msau\_big$ has the same configuration except for the number of residual blocks is 5. The output of the first block is used for intermediate supervision. 

\textbf{Baselines}.
% Given the steady performance and efficiency, the U-Net architecture is adopted as our baselines. 
In our experiment, we used two U-Net variants with dilated convolution \cite{atrous_conv_Papandreou2015, Dilated_Conv_Chena} and residual bottleneck in the downsampling path, which is similar to the encoder backbone of our architecture.
The first variant has $n_{downsampling}=5$, $res\_depth=2$ and $C=16$. 
% The first variant has 5 residual blocks with the residual depth $res\_depth=2$ and the base feature depth of 16. 
% We dubbed this model as $unet\_small$. 
% The second variant has 6 residual blocks and the $res\_depth$ of 3, called $unet\_big$. 
The second variant has $n_{downsampling}=6$, $res\_depth=3$. %, called $unet\_big$. 
We abbreviate them as $unet\_small$ and $unet\_big$, respectively. 
These two networks are very strong baselines due to the proven effectiveness of their sub-components. 
Additionally, the same architecture was also presented in \cite{Katti2018}, which allows us to make a direct comparison to one of the current state-of-the-art methods. 
%which serves the purpose of comparison with our method. 
Table \ref{table:param} summarized the number of parameters of our network and the baselines. 
As will be shown shortly, although our $msau\_small$ uses 40\% less parameters, it is still able to outperform $unet\_big$ by a large margin in term of accuracy. 

\textbf{Metrics}.
To compare the performance across models, we use three metrics: \textit{mean Intersection-over-Union} (mIOU), \textit{mean pixel accuracy} (pix acc) and the \textit{box F1-score} (F1-score). 
%While the previous metrics are calculated by the pixel-level segmentation output, 
We note that we leveraged a modified metric, so-called box F1-score, to evaluate the correctness of the predicted bounding box for each key field. % in the document. 
The bounding box of each field is constructed from the connected component analysis on the output segmentation mask. 
A predicted box is marked as a correct detection if its IoU ratio with one of the ground-truth boxes is larger than a certain threshold $0.8$. 
This metric is suitable for testing the model on practical OCR-ed document where the ground-truth mask may slightly misalign with the generated char-grid from the OCR text.

\begin{table}[h]
\centering
\footnotesize
\begin{minipage}{0.32 \textwidth}
\caption{Number of \hspace{\textwidth} trainable parameters}
\begin{tabular}{|l | c |} 
 \hline
 Model & Number of \\ [0.9ex]
 & parameters \\
 \hline\hline
 $unet\_small$ & $6.7 \times 10^5$\\ 
 $unet\_big$ & $10.6 \times 10^5$  \\
 \hline
 $msau$ (ours) & $6.6 \times 10^5$ \\
 $msau\_big$ (ours) & $10.5 \times 10^5$ \\
 \hline
\end{tabular}
\label{table:param}
\end{minipage}
\begin{minipage}{0.45\textwidth}
\caption{Performance comparison \hspace{\textwidth} between ground-truth and OCR text}
\begin{tabular}{|l | l | c | c |} 
 \hline
 Dataset & Model & F1-score & F1-score \\ [0.9ex] 
 & &  (GT text) & (OCR) \\
 \hline\hline
 Invoice & $unet\_big$ & $87.3$ & $79.7$ ($-8\%$)\\
 & $msau\_big$ (ours) & $96.0$ & $91.2$ ($-5\%$)\\
 \hline
 Insurance & $unet\_big^{}$ & $89.9$ & $83.4$ ($-7\%$) \\
 & $msau\_big$ (ours) & $96.1$ & $92.4$ ($-4\%$)\\
 \hline
\end{tabular}
\label{table:overall_ocr}
\end{minipage}
\end{table}

\subsection{Results}

\textbf{Overall results}.
We first compare the performance of our $msau$ models to the baseline architectures. 
Table \ref{table:overall_invoice} shows the results of all models that have been deployed on the Japanese invoices testset. 
As we can see, the MSAU-nets surpassed the baseline U-Net models in all metrics by a significant margin. 
The base $msau$ model improved the $mIOU$ score by $7\%$ compare to the $unet\_big$, with only 60\% the number of parameters. 
The $msau\_big$ further increased this gap to 9\%, leading to an remarkable 40\% error reduction. 
This demonstrates the effectiveness and efficiency of our architectures. 
$F1$-$score$ is also largely improved, with a 9\% increase from $unet\_big$ to $msau\_big$ and 7\% increase from $unet\_big$ to $msau$. 

% \begin{table}[h]
% \centering
% \footnotesize
% \begin{tabular}{|l | c c c|} 
%  \hline
%  Model & mIOU & pix acc & F1-score \\ [0.9ex] 
%  \hline\hline
%  $unet\_small$ & $74.4$ & $79.2$ & $83.2$ \\ 
%  $unet\_big$ & $78.0$ & $85.7$ & $87.3$ \\
%  \hline
%  $msau$ (ours) & $85.5$ & $91.1$ & $94.6$ \\
%  $msau\_big$ (ours) & $\textbf{87.2}$ & $\textbf{92.5}$ & $\textbf{96.0}$ \\
%  \hline
% \end{tabular}
% \caption{Performance comparison on the Japanese invoices dataset}
% \label{table:overall_toyota}
% \end{table}

\begin{table}[h]
\centering
\footnotesize
    \caption{Comparison on Japanese invoices (left) and medical receipts (right) datasets}
    \begin{minipage}{.45\linewidth}
        \begin{tabular}{|l | c c c|} 
         \hline
         Model & mIOU & pix acc & F1-score \\ [0.9ex] 
         \hline\hline
         $unet\_small$ & $74.4$ & $79.2$ & $83.2$ \\ 
         $unet\_big$ & $78.0$ & $85.7$ & $87.3$ \\
         \hline
         $msau$ (ours) & $85.5$ & $91.1$ & $94.6$ \\
         $msau\_big$ (ours) & $\textbf{87.2}$ & $\textbf{92.5}$ & $\textbf{96.0}$ \\
         \hline
        \end{tabular}
        %\caption{a}
        \label{table:overall_invoice}
    \end{minipage}
    \begin{minipage}{.45\linewidth}
        \begin{tabular}{|l | c c c|} 
         \hline
         Model & mIOU & pix acc & F1-score \\ [0.9ex] 
         \hline\hline
         $unet\_small$ & $78.9$ & $83.3$ & $86.2$ \\ 
         $unet\_big$ & $81.7$ & $86.5$ & $89.9$ \\
         \hline
         $msau$ (ours) & $86.4$ & $91.8$ & $95.0$ \\
         $msau\_big$ (ours) & $\textbf{89.1}$ & $\textbf{93.3}$ & $\textbf{96.1}$ \\
         \hline
        \end{tabular}
        %\caption{b}
        \label{table:overall_myl}
    \end{minipage}
\end{table}

% The  the results for the Insurance medical receipts dataset are summarized in Table \ref{table:overall_myl}. 
Similar achievements are expected while testing on the Insurance medical receipts datasets (Table \ref{table:overall_myl}).
The MSAU models continue to outperform the baseline U-Nets consistently. 
The $msau\_big$ model leads to an obvious 8\% improvement in $mIOU$ from the $unet\_big$. 
The base $msau$ also demonstrated good parameter efficiency by delivering better performance compared to $unet\_big$ (5\%) in both $mIOU$ and $F1$-$score$. 

We further evaluate our models on the two datasets with output text from an OCR engine (Tesseract \cite{tesseract_Smith2007}). 
%Results are shown in table \ref{table:overall_ocr}. 
As can be seen in Table \ref{table:overall_ocr},  
the $msau\_big$ model shows its robustness to OCR errors with 5\% and 4\% drops in $F1$-$score$, while the baseline $unet\_big$ has 8\% and 7\% drops respectively. 
Furthermore, the gap between $unet\_big$ and $msau\_big$ is enlarged to 9$\sim$10\% in the OCR-ed documents. 
%This proves the advantage of our model in a real document processing pipeline, where the errors from OCR step are inevitable.  
This shows that our method is able to detect the errors produced by OCR step by a successful rate.

\begin{figure}[h]
    \centering
    \subfigure[Performance with different OCR error rates $e$ on the Japanese Invoice dataset]{
        \includegraphics[width=0.45\textwidth]{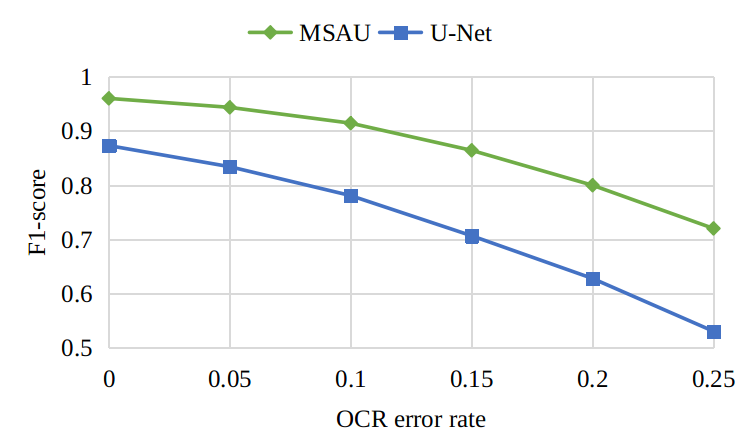}
        \label{fig:msau_ocr}
    }
    \subfigure[Performance with different train/val ratios on the Japanese Invoice dataset]{
        \includegraphics[width=0.45\textwidth]{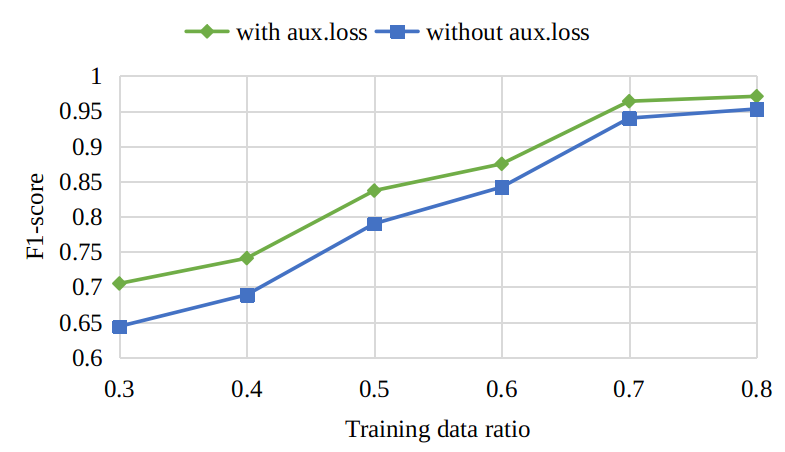}
        \label{fig:train_ratio}
    }
    \caption{Performance comparison in two different scenarios.} %constrained condition}
\end{figure}

For a more comprehensive comparison, we show our model results across different levels of synthetic OCR errors (randomly replacing/removing character with probability $e$) in Figure \ref{fig:msau_ocr}. 
The impact of OCR errors is less severe on our MSAU-net as the proposed model still achieve more than 70\% $F1$-$score$ in an extreme condition where character error rate is as high as 25\%. 

\noindent
\textbf{Ablation study of sub-components}.
In this part, we perform a step-wise decomposition on our model to study the effect of each sub-component. 
Experiment is conducted using the $unet\_big$ and $msau\_big$ baseline on the Japanese Invoice dataset. 
The performance results are shown in Table \ref{table:ablation}.

\begin{table}[h]
\centering
\caption{Performance comparison with different model configurations. \small{Abbreviations: NL - \textit{Non-local self-attention}, MS - \textit{Multi-Stage learning}, BC - \textit{Box Convolution}, CUNet - \textit{Coupled U-Net} }}
\footnotesize
\begin{tabular}{| l | l | c | c c|} 
 \hline
 \# & Model & Params ($\times10^5$) & mIOU & F1-score \\ [0.9ex] 
 \hline\hline
 1 & UNet ($unet\_big$) & $10.6$ & $78.0$ & $87.3$ \\ 
 2 & UNet+NL & $10.5$ & $79.2$ & $88.1$ \\
 3 & UNet+NL+BC & $10.8$ & $\textbf{80.7}$ & $\textbf{91.2}$  \\
 \hline
 4 & CUNet & $10.3$ & $79.5$ & $89.6$ \\
 5 & CUNet+MS & $10.3$ & $82.3$ & $92.5$ \\
 6 & CUNet+NL & $10.5$ & $81.5$ & $91.3$ \\
 7 & CUNet+MS+BC & $10.3$ & $86.2$ & $95.1$  \\
 8 & CUNet+NL+BC & $10.5$ & $84.6$ & $94.6$  \\
 9 & CUNet+NL+BC+MS ($msau\_big$) & $10.5$ & $\textbf{87.2}$ & $\textbf{96.0}$  \\
 \hline
\end{tabular}
\label{table:ablation}
\end{table}

\begin{figure}[h]
\centering
\includegraphics[width=0.4\textwidth]{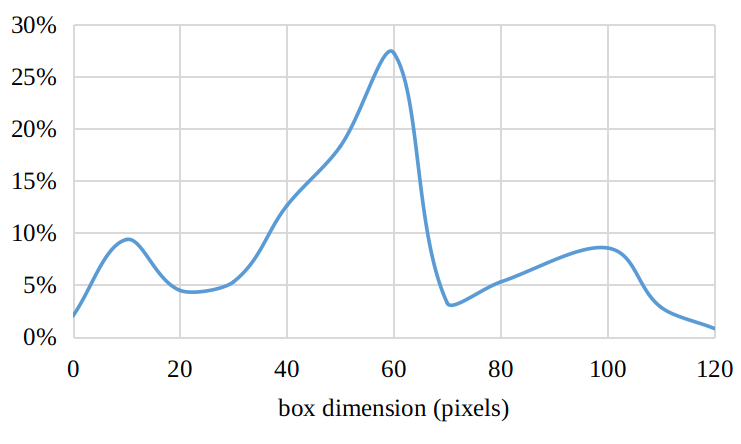}
\caption{Distribution of filter sizes in the box convolution layer. \small{\textit{The horizontal axis shows the max dimension of learned box filters)}}}
\label{fig:box_dist}
\end{figure}

As we can see, the base CU-Net model (4) improves the mIOU score by 2\% compared to the base U-Net (1). 
It also surpasses the UNet+NL (2) configuration. 
The Non-local (self-attention) block helps the base model gained 1\% and 2\% in $mIOU$, respectively, in cases of the  U-Net (2) and CU-Net (6) baseline.
The CU-Net configuration gained much greater improvement in comparison with the base U-Net when adding the box convolution (3\% and 1\% $mIOU$ in (3) and (8)), respectively. 
This demonstrates the effectiveness of the component, especially when combined with the CU-Net for modeling spatial relation in the char-grid. 
% Notably, with the multi-stage training (5), the box convolution achieved 4\% $mIOU$ increase from (7), which is impressive considering this is the same gap between the $unet\_big$ and $unet\_small$ model with substantial difference in parameter count. 
Notably, with the multi-stage training (5) plus the box convolution, configuration (7) improves the overall $mIOU$ by 4\%. 
This is a significant achievement because in order to reach the same number, $unet\_small$ has to transform to $unet\_big$ with more 40\% parameters. 
The effect of box convolution can be further analyzed in the Figure \ref{fig:box_dist}, in which we can see that the improved result come from the larger receptive field by expanding the box convolution filter dimension. 
%The receptive field is substantially greater than the normal convolution layer.

\noindent
\textbf{Ablation study of multi-stage training}. Table \ref{table:ablation} shows that the multi-stage training (9) increased the $mIOU$ by a significant 3\%, compared to the CUNet+NL+BC (8) configuration. 
This observation also holds in the base CU-Net model with 3\% improvement from (4) to (5). 
We can see that the multi-stage training greatly helps to gain the performance while introducing no additional complexity to the model. 
One explanation is that the separation of textual and spatial feature learning in the auxiliary loss can encourage each sub-component to focus on certain task, hence better generalization. 
Note that the multi-stage paradigm can not be applied to the U-Net baseline due to the single encoder-decoder design. 
The best CU-Net configuration with multi-stage greatly outperformed the best U-Net model by more than 6.5\% $mIOU$. 
Lastly, we found that the multi-stage training also helps increase the robustness of the model in limited training data scenarios. 
Figure \ref{fig:train_ratio} shows the impact of different train/val ratios on the model performance. 
It can be seen that the auxiliary loss significantly boost the performance in all cases. 
More importantly, the improvement is more noticeable in the case of less training data. 

\begin{figure}[h]
\centering
\includegraphics[width=0.85\textwidth]{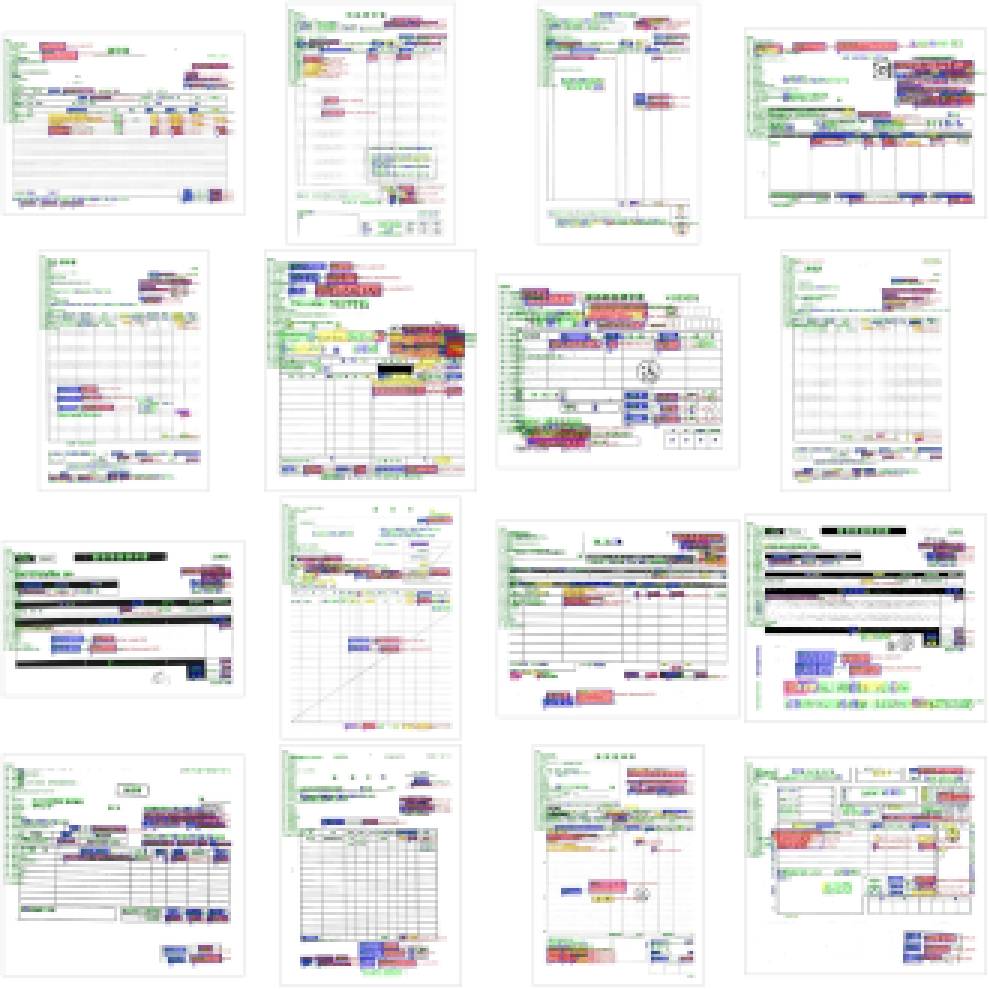}
\caption{Sample outputs on the Japanese invoice dataset. \small{\textit{Segmented key mask is presented in blue, value mask is presented in red}}}
\label{fig:invoice_output}
\end{figure}

\noindent
\textbf{Qualitative result}. We provide a qualitative visualization of our algorithm in Figure \ref{fig:invoice_output}. More results are provided in the \href{https://drive.google.com/file/d/1FFL1cjr65XcVMaeoDn5cx-uoOWOMSqT8/view?usp=sharing}{supplementary}, and we encourage the reader to view them at full size on a screen. 
\section{Conclusion}
We have presented a novel deep neural network, Multi-Stage Attentional U-Net (MSAU), for end-to-end information extraction on the 2D char-grid representation of the document. 
We incorporate attention mechanism, box convolution with the multi-stage encoder-decoder architecture to handle complex textual and spatial relation in the char-grid. 
Moreover, we propose a multi-task training scheme to improve the model performance on practically challenging scenarios. 
Experiments on two benchmark datasets have demonstrated the effectiveness of our approach. 
In the future work, we wish to open-source our datasets and provide the proposed methods as new baselines to promote the current active research in document analysis area. \\
\textbf{Acknowledgement}. The authors wish to offer a sincere thanks to Cinnamon AI Labs for providing the required resources to complete this work. We also express our deep gratitude to Tran Minh Quan (lotus@cinnamon.is) and Nghiem Nguyen Viet Dung (henry@cinnamon.is) for their valuable aids in the final preparation of the paper. 
\bibliography{library}

% \section*{Supplementary Figures}
% \input{supplemental.tex}
\end{document}